%% file: main.tex
\pdfoutput=1

\documentclass[11pt]{article}

\usepackage[]{acl}

\usepackage{times}
\usepackage{latexsym}
\usepackage{CJKutf8}

\usepackage[utf8]{inputenc}
\usepackage{graphicx}

\usepackage{inconsolata}

\usepackage[T1]{fontenc}

\usepackage[utf8]{inputenc}

\usepackage{microtype}

\usepackage{defs}
\usepackage{defs2}
\usepackage{defs3}
\usepackage{clues_defs}
\usepackage{attention}
\usepackage{math-common}
\usepackage{makecell}

\usepackage{caption}
\usepackage{subcaption}
\usepackage{xcolor}
\usepackage{inconsolata}
\definecolor{chromeyellow}{rgb}{1.0, 0.65, 0.0}

\newcommand{\nop}[1]{}

\title{Revisiting Sample Size Determination in Natural Language Understanding}

\author{Ernie Chang$^{\dag\footnotemark[1]}$ , Muhammad Hassan Rashid$^{\ddagger\footnotemark[1]}$ , Pin-Jie Lin$^{\ddagger\footnotemark[1]}$ ,\\ {\bf Changsheng Zhao$^{\dag}$}, {\bf Vera Demberg$^{\ddagger}$}, {\bf Yangyang Shi$^{\dag}$} and {\bf Vikas Chandra$^{\dag}$} \\
    $\dag$Reality Labs, Meta Inc. \\
    $\ddagger$Saarland Informatics Campus, Saarland University, Germany \\
    \texttt{\{erniecyc, cszhao, yyshi, vchandra\}@meta.com} \\
    \texttt{hassanrashid725@gmail.com} \\
    \texttt{pinjie@lst.uni-saarland.de} \\
    \texttt{vera@coli.uni-saarland.de}}
\begin{document}
\maketitle

\newcommand\blfootnote[1]{%
  \begingroup
  \renewcommand\thefootnote{}\footnote{#1}%
  \addtocounter{footnote}{-1}%
  \endgroup
}
\blfootnote{$\ast$ These authors contributed equally to this work.}

\begin{abstract}

\input{sections/0_abstract}

\end{abstract}

\input{sections/1_intro}

\input{sections/2_related}

\input{sections/3_approach}

\input{sections/4_experiments}

\input{sections/5_analysis}
\input{sections/conclusion}

\input{sections/6_limitations}

\bibliography{anthology,sample_size,datasets,custom}
\bibliographystyle{acl_natbib}

\appendix
\clearpage

\input{sections/appendix.tex}

\end{document}

%% file: sections/0_abstract.tex
Knowing exactly how many data points need to be labeled to achieve a certain model performance is a hugely beneficial step towards reducing the overall budgets for annotation.
It pertains to both active learning and traditional data annotation, and is particularly beneficial for low resource scenarios. 
Nevertheless, it remains a largely under-explored area of research in NLP.
We therefore explored various techniques for estimating the training sample size necessary to achieve a targeted performance value. We derived a simple yet effective approach to predict the maximum achievable model performance based on small amount of training samples -- which serves as an early indicator during data annotation for data quality and sample size determination.
We performed ablation studies on four language understanding tasks, and showed that the proposed approach allows us to forecast model performance within a small margin of mean absolute error ($\backsim 0.9$\%) with only 10\% data\footnote{Our code is available at: \url{https://github.com/pjlintw/sample-size}.}.


%% file: sections/1_intro.tex
\section{Introduction}

Labeled data play an important role in creating performant machine learning models, which makes data annotation a fundamental process for any natural language application pipeline~\cite{lewis1994heterogeneous,chang-etal-2020-dart}.
Recent work has sought to reduce the annotation costs through the use of active learning~\cite{ducoffe2018adversarial,margatina2021active} and data sampling~\cite{sener2018active,coleman2019selection,pmlr-v139-killamsetty21a,killamsetty2021retrieve,chang-etal-2021-training}.
Indeed, these approaches are shown to be effective in identifying or constructing data subsets needed to achieve a competitive model performance. 
For instance, the active learning paradigm adds new data iteratively to the existing set before model retraining~\cite{agarwal2020contextual,margatina2021active}, improving upon the traditional human annotation pipeline that obtains the entire labeled set all at once.

Nevertheless, the data labeling process typically annotates as much data as the annotation budget permits, or by clearly defined stopping criteria to terminate the labeling process.
Unfortunately, this is usually challenging as annotators do not have the knowledge of the effect of added labels to model performance nor how much more data is needed to arrive at the desired model generalizability ~\cite{DBLP:journals/corr/abs-2012-10630}.
The stopping condition is in fact tied to the quality of data samples w.r.t. model parameters~\cite{hu2021model}, 
which influences the effective sample size\footnote{It is the size of datasets which could have been achieved by an effective unweighted random sample~\cite{Guo2022-vb}.}, 
and it is then beneficial to obtain an approximation of the expected performance ~\cite{VLACHOS2008295,olsson-tomanek-2009-intrinsic,10.1145/1753783.1753784,pmlr-v108-ishibashi20a}.
Therefore, knowing the approximate amount of training data needed for this particular performance would serve as an useful knowledge not only for deciding when to stop adding labeled data, but also as an early indication for the data quality. 
For instance, by having early label quality signals, we can decide between two different types of annotation, or even between two pools of annotators with different expertise.
 
To this end, we explored the relationship between \emph{data sample size} and \emph{model performance} in the context of language understanding via learning curve modeling, which defines model performance as a function of dataset sizes.
By modeling this relationship in low resource settings, we obtain useful early signals with approximated accuracies for any given the labeled set, which can provide an idea for the sample size and data quality~\cite{fredrik2009activeLearning,rosa2012predicting}.
Previous studies have shown that nonlinear weighted curve fitting methods such as inverse power laws or exponential functions can provide decent approximations of the empirical predictive performances~\cite{pmlr-vR2-frey99a,rosa2012predicting}. 
We thus put forward an ensemble of these functions which we showed to display a consistently highly correlated behavior across four language understanding benchmarks and with as little as 10\% of the entire training set. 
This work makes the following contributions:
\begin{enumerate}
    \item We revisit the task of sample size determination in four natural language understanding benchmarks and empirically explore the correlation strengths of several successful techniques. 
    \item Based on our findings, we propose an \textsc{Ensemble} function and demonstrated across several benchmarks and low resource settings that the ensemble function is consistently providing a high correlation with the empirical learning curve plots.
\end{enumerate}

%% file: sections/2_related.tex
\section{Background}

Our method is a sample size determination technique that helps to design annotation projects by determining the necessary sample size. Previous methods have focused on identifying the sample size required to reach a specific target performance, such as a high correlation coefficient~\cite{Beal1989-ss,Stalbovskaya2007-dj,Beal1989-ss}, which often involves predicting the sample size necessary for a classifier to attain a specific accuracy level~\cite{Fukunaga1989-hn}. There are two main approaches for predicting the sample size needed to achieve a particular classifier performance: (1) \citet{Dobbin2008-fi} present a model-based method for predicting the number of samples required for classifying microarray data. (2) A more general approach involves fitting a classifier's learning curve to inverse power law models~\cite{rosa2012predicting}. Examples of this approach include algorithms proposed by \citet{mukherjee2003predicting,boonyanunta2004predicting,last2007predicting}.

%% file: sections/3_approach.tex

\section{The Approach}

\paragraph{Learning Curve Modeling.}

A learning curve is a graphical representation of how a classifier's performance changes as the size of the training set increases. 
The curve typically has three sections: an initial section where performance improves rapidly with increasing training set size, a middle section where the rate of improvement slows down, and a final section where the classifier reaches its maximum performance and further increases in training set size do not lead to significant improvements. 
This relationship can be quantified using a set of data points, each of which represents the expected performance of the classifier $E_{acc}$ on a particular training set size $D_{k}$. 
These data points can be plotted to create the learning curve, which can help to understand the behavior of the classifier and inform decision-making about how much training data is needed to achieve a desired performance level.

\paragraph{Task Description.}

Given a downstream classification task with $N_{total}$ data points, a learning curve model $F$ predicts the expected performance $E_{acc}$ when a classifier trained on the an observed range of training set size ($D_k$; $k>=N$). The empirical learning curve is assessed by the parametric models for the learning algorithm performance extrapolation.
In our settings, we set $k << N_{total}$ to simulate practical settings, where few data points consisting of $(E_{acc}, D_K)$ are to be obtained.

\paragraph{Types of Extrapolations.}

Here, we study different forms of learning curve models with few learnable parameters that have been proven as simple yet effective. 
The simplest type of learning curve model \emph{exponential function} (\textsc{Exp}) only introduces two parameters $a$ and $b$ to fit the exponent behavior of learning curve~\cite{pmlr-vR2-frey99a}. 
The second form, \emph{Inverse Power Law function} (\textsc{Inverse}), fits the inverse power law ~\cite{rosa2012predicting} and has three parameters.
The third form uses a function from the power law family -- Power4 function (\textsc{Pow4}) ~\cite{kolachina-etal-2012-prediction} with four parameters. 
Lastly, we propose to combine all functions into one (\textsc{Ensemble}) so that it has all their characteristics in order to make it more robust across benchmarks.
Table~\ref{table:functions} shows the formulae of our investigated extrapolating functions.

\begin{table}[!th]
\centering
\resizebox{0.95\columnwidth}{!}{%
\centering
\begin{tabular}{lc}
\Xhline{2\arrayrulewidth}

\textsc{Extrapolating Functions} & \textsc{Formula}   \\
\hline 
\textsc{exp} (A)  & $a \cdot N^{b}$  \\ 
\textsc{inverse} (B) & $(1-a) - b \cdot N^c$  \\
\textsc{pow4} (C) & $a - (b \cdot N+c)^{-d}$  \\
\textsc{Ensemble (A+B+C)} &  $-$ \\
\Xhline{2\arrayrulewidth}
\end{tabular}}
\caption{\small Overview of extrapolating functions}
\label{table:functions}
\end{table}

%% file: sections/4_experiments.tex
\section{Experimental Settings}

We study four NLU tasks: 
(1) \textsc{IMDb} \cite{maas2011learning} is a binary classification dataset ($25$K/--/$25$K)\footnote{Expressed in the order (train/dev/test).} where model predicts the sentiment (positive/negative) for movie reviews from IMDB; 
(2) \textsc{SST2} \cite{socher2013recursive} is also a sentiment classification datatset ($67$K/$0.8$K/$1.8$K) containing reviews of different movies and since the model predicts if the review is positive or negative, it also falls in the category of binary classification; 
(3) \textsc{AG NEWS} is a multi-class classification dataset ($120$K/--/$7.6$K) containing texts from different news where the model predicts whether the news text is about sports, science/technology, world or business from the four different classes.
We also consider one other multi-class classification task, (4) \textsc{DBpedia} dataset ($560$K/--/$70$K) , since it could help us in testing the robustness of the methods used in our experiments.

\paragraph{Configs.} To investigate how changes in data size affect the predictiveness of the learning curves, under the assumption that the model structure and settings remain unchanged, we perform all experiments using a transformer model \cite{vaswani2017attention} and average the results over 3 initialization runs. The embedding and hidden layer dimensions are 1000 and 1820; and we use a 6-layer encoder with 4 multi-heads, and the dropout is 0.2. 
To find the parameters of learning curve models, we consider {unweighted} and for the gradient descent and non-linear least squares optimizers.
The Adam algorithm~\cite{kingma2014adam} was used as the optimizer with learning rate of 1e-5 and ReLU was used as the activation function. 
The cross-entropy objective was used for all classification benchmarks, and we select the models using loss values.
Finally, we chose a batch size of 8 with 200 number of epochs. 

\paragraph{Evaluation.}

We use the aforementioned functions: \textsc{Exp}, \textsc{Inverse}, \textsc{Pow4} and \textsc{Ensemble} for fitting the empirical learning curve.
For each dataset, we select training set sizes ranging from $1$\% to $10$\% data sizes at an interval of $1$\%. 
The learning curve testsets were created with the data splits in the range $[55, 100]$ at $5$\% interval by training the classifier, and obtaining the testset\footnote{Here, we make the distinction
between testset for learning curve and the original testset split.} performance for each corresponding data split. 
Therefore, we collect the accuracies against different sample sizes and
report the mean absolute error (MAE) as the evaluation metric for learning curve modeling.




\section{Results and Analysis}

We present results of ensemble method for learning curve modeling on the NLU benchmarks.

%% file: sections/5_analysis.tex
\subsection{Main Results}

\begin{figure*}[!t]
\centering
    \begin{subfigure}{0.5\textwidth}
        \centering
        \includegraphics[width=\textwidth]{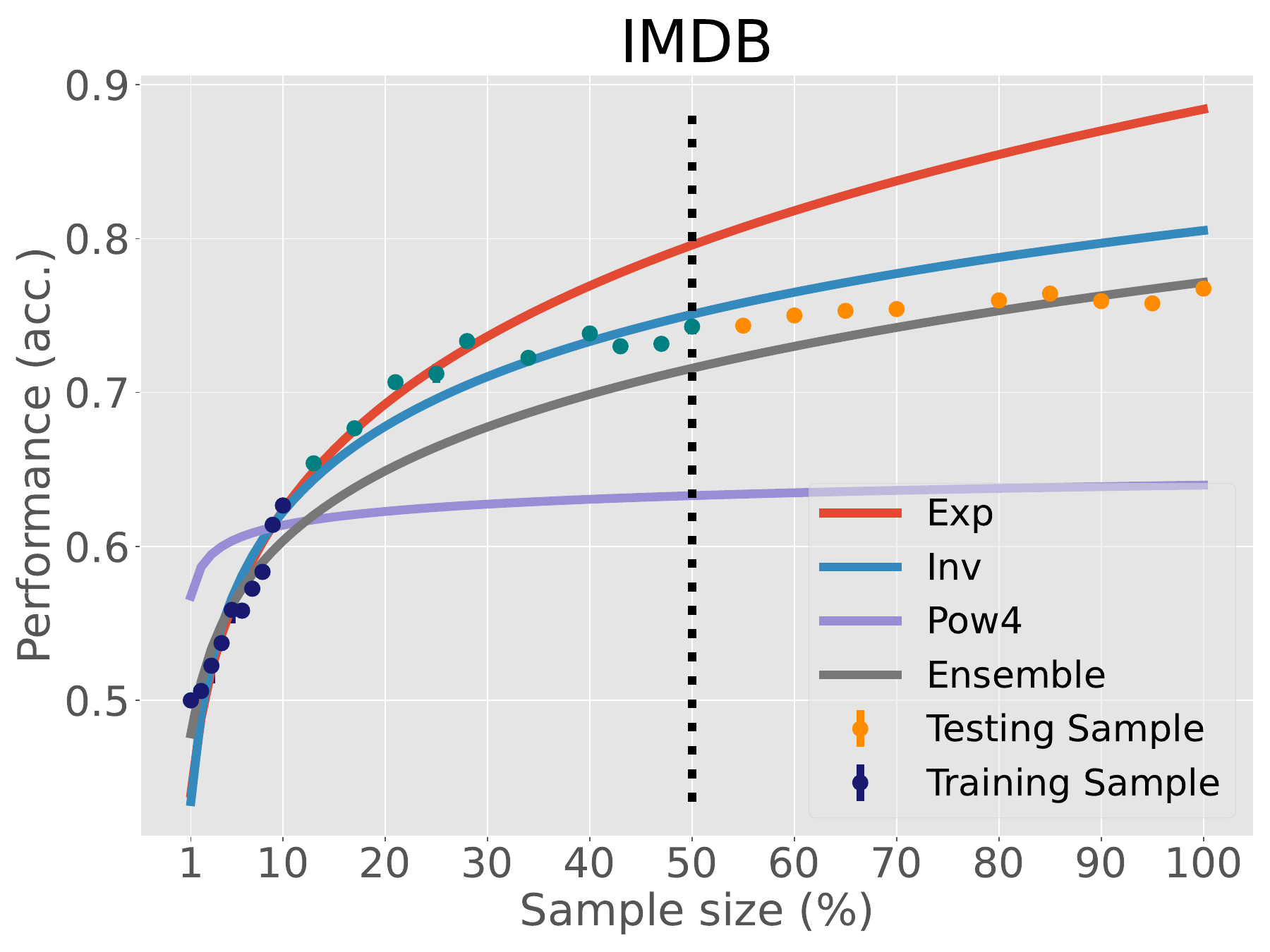}
    \end{subfigure}%
    \begin{subfigure}{0.5\textwidth}
        \centering
        \includegraphics[width=\textwidth]{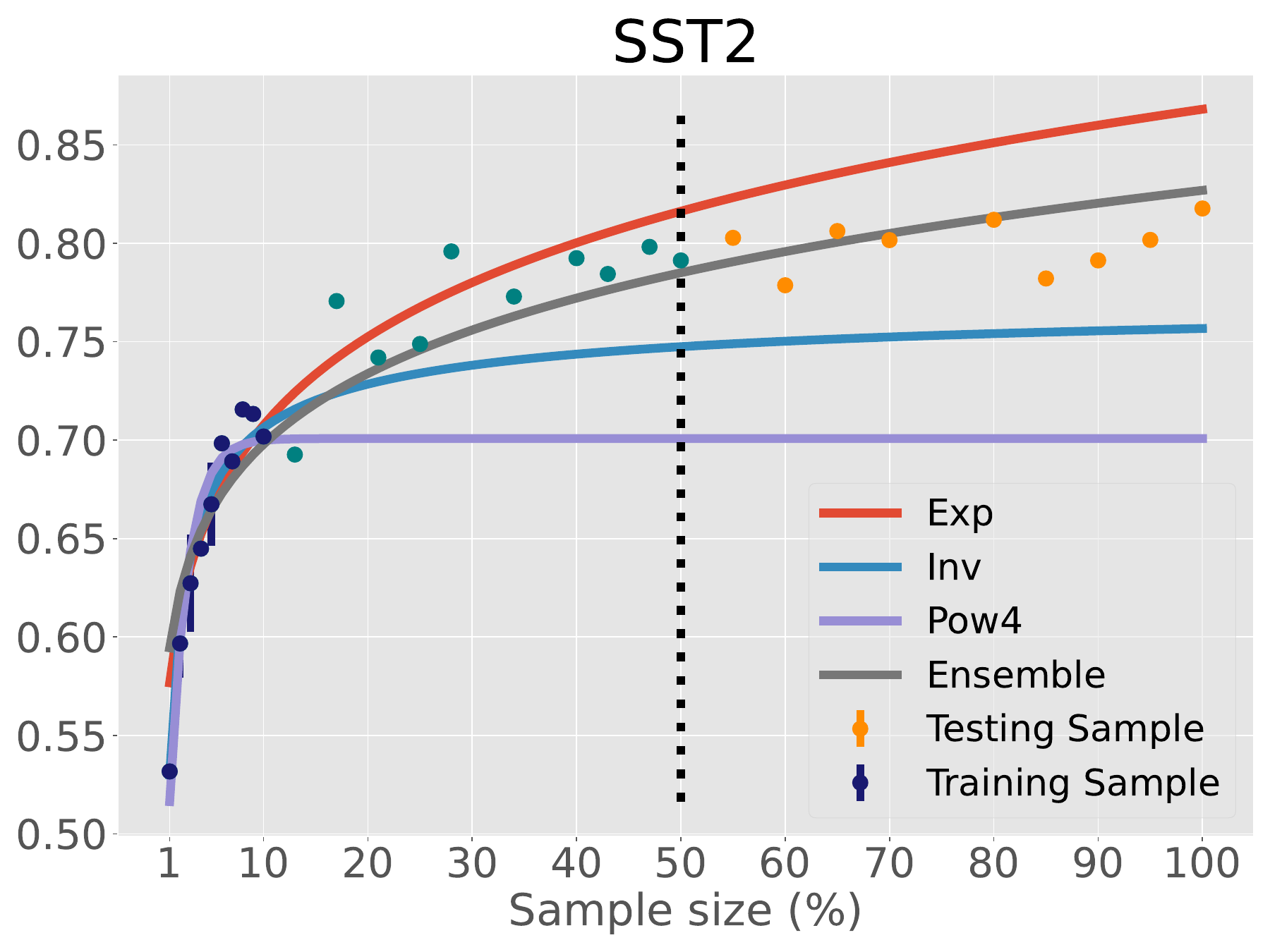}
    \end{subfigure}%
    \caption{\small Learning curves on $10\%$ sample size of IMDB and SST2 datasets. We plot learning curves using the exponential (Exp), inverse power law (Inv), power4 (Pow4) function, and the combination of the aforementioned forms (Ensemble). The learning curves only fit on $10\%$ training sample ({\color{blue}blue}) and generalize on the unseen data sizes. We evaluate the learning curves on the testing sample ({\color{chromeyellow}yellow}). Data sizes from $10\%$ to $50\%$ ({\color{teal}teal}) are neither used in training nor testing.}
    \label{fig:learning_curve_10_pct}
\end{figure*}

Figure~\ref{fig:learning_curve_10_pct}
demonstrates that by using only 10\% of the data for learning curve modeling, \textsc{Ensemble} is able to effectively predict model performance within a 0.9\% margin of the actual model performance. 
Moreover, we observe the same trend across all four benchmarks consisting of different training set sizes (i.e. ranging from 25K to 250K) and varying number of classification classes (i.e. ranging from 2 to 14), see the appendix \ref{appendix_detailed_results} for remaining figures.
Our result shows that the proposed approach is not confined by the classification types and sample sizes.

Table~\ref{table:main} shows the saturated points of the learning curve when the performance improvement is less than a threshold $\alpha=0.2$ -- we found that the predicted performance with only $19$\% data is within $2.44$ accuracy points from the trained model performance for \textsc{IMDb}.
Another key observation is that the size (\%) needed to predict a low L1 distance increases as the number of classification classes goes up, which indicates that task difficulty does influence the ease of extrapolation.
An example is that \textsc{AG News} requires up to $51$\% to predict a low L1 distance.
Next, we perform further ablation studies to investigate the effect of sample size, types of non-linear functions used, or the effect of data weighting.

\begin{table}[ht]
\centering
\resizebox{\columnwidth}{!}{
\centering
\begin{tabular}{lccccc}
\Xhline{2\arrayrulewidth}
\textsc{Benchmark} & \textsc{Cls (\#N)} & \textsc{Size (\%)} & \textsc{Size (\#N)}  & \textsc{L1}$\downarrow$  & \textsc{100\%} \\\hline
\multicolumn{6}{c}{\textsc{$\alpha=0.2$}}\\\hline
\textsc{IMDb} & $2$ & $36$\% &  $6,300$ & $2.44$ & $17$K\\ 
\textsc{SST2} & $2$ & $19$\%  & $8,958$ & $5.57$ & $47$K \\
\textsc{AG News} & $4$ & $51$\% & $42,840$ & $2.6$ & $84$K \\
\textsc{DBpedia} & $14$ & $51$\% & $199,920$ & $2.39$ & $392$K \\
\Xhline{2\arrayrulewidth}
\end{tabular}}
\caption{\small \textsc{Cls (\#N)} stands for the number of classes, \textsc{Size (\%)} for the percentages of the data size for the learning curve modeling. \textsc{Size (\#N)} is the number of the corresponding data size, \textsc{L1} is the L1 distance between the accuracy of models using all the training data and the estimated accuracy based on the saturated point. \textsc{100\%} specifies all training samples for learning curve.}
\label{table:main}
\end{table}
	
\subsection{Ablation Study}
\paragraph{Effect of sample size.} 

In Figure~\ref{fig:learning_curve_10_pct}, we study the correlation between sample sizes and the absolute mean error between the learning curve model and empirical model performance trend. Surprisingly, we discovered by having more samples does not necessarily help with modeling a better learning curve\footnote{We showed this result in the Appendix \ref{sec:sample_sizes}.}, and that with only $10$\% data to build the $(D_k, E_{acc})$ data points is sufficient to obtain rather small errors across all four benchmarks.

\paragraph{Types of learning curve functions.}

We are also interested in seeing how each of the non-linear learning curve function fare against each other in simpler settings. 
To this end, we used up to 10\% data to model the learning curves and obtained their respective mean absolute error values. 
In Figure~\ref{fig:learning_curve_10_pct}, we present this comparison where we showed that on \textsc{IMDb} and \textsc{SST2}, the  \textsc{Ensemble} function consistently fit best against the empirical data. 
We observed a similar trend across other benchmark \textsc{DBpedia} with the exception of \textsc{AG NEWS}.
We placed the plot for \textsc{AG NEWS} in appendix ~\ref{lr_on_ag_news}. 

\paragraph{Influence of data weighting.} 
Previous work~\cite{paul2021deep,Guo2022-vb} has found that not all data points are equally important in terms of curve fitting. 
In fact, data points at a later phase corresponding to more samples are to be given more weight compared to earlier points. 
We thus investigate this phenomenon in the context of our benchmark, and we observed this to be true anecdotally. The detailed result can be found in Appendix \ref{sec:data_weighting}. The reason for this is that the more data samples there are, the more closely they resemble the entire training set, and this makes their signals a better estimation of a point on the actual learning curve. 
Another perspective is that the more data samples are used, the less the effect of random sampling on the performance, which affects model performance in extremely low resource scenarios.

\begin{table}[ht]
\centering
\resizebox{\columnwidth}{!}{
\centering
\begin{tabular}{lcc}
\Xhline{2\arrayrulewidth}
\multirow{2}{*}{\textsc{Function Type}} & \multicolumn{2}{c}{\textsc{Non-Linear Least Squares}} \\
\cline{2-3}
& \textsc{Unweighted} & \textsc{Weighted} \\
\hline
\textsc{Exp}  & $0.0417$ &  $\textbf{0.0244}$  \\ 
\textsc{Inv} & $0.00777$  & $\textbf{0.00442}$ \\
\textsc{Pow4}  & $0.00795$ & $0.00795$ \\

\Xhline{2\arrayrulewidth}
\end{tabular}}
\caption{\small Better curve fitting when weighting data points at latter phase. We examine the effectiveness of weighting data size on the exponential (\textsc{Exp}), inverse power law (\textsc{Inv}), power4 (\textsc{Pow4}) function using non-linear least squares method. The learning curves fit on 5\%, 10\%, 25\% and 50\% data sizes of IMDB and is evaluated on testing sample with mean absolute error (MAE).}
\label{table:weighting_data_size}
\end{table}

%% file: sections/conclusion.tex
\vspace{-0.5cm}

\section{Conclusions and Future Works}

In this work, we investigated techniques for estimating the amount of training data needed to achieve a target performance in four natural language understanding benchmarks. 
We demonstrated that our approach allows for accurate prediction of model performance using only a small portion of the data, which can be useful in scenarios with limited resources. 
Nevertheless, we also recognize the limitation in our current study. 
For instance, we did not explore sampling techniques other than random sampling; while recent works~\cite{yuan-etal-2020-cold,paul2021deep,Guo2022-vb} have shown promising directions in data sampling that outperforms random sampling. 
Another interesting direction is to explore the model architecture's influence on generalizability, and thus the learning curve, which we left for future works.

%% file: sections/6_limitations.tex
\section*{Limitations}

While the effectiveness of the expressive learning curve in settings with limited data has been demonstrated, it is uncertain if this success can be replicated in more complex natural language understanding tasks, such as question answering or tasks that involve a large amount of data.
Furthermore, it is assumed that all data samples have the same impact on the model's performance. However, the actual performance of the model may vary based on the method used to select the data or the specific set of tasks being performed, e.g., coreset selection. Similarly, the quality of the labels used for the data can also play a significant role in predicting the performance of the model. Overall, we plan to further investigate these questions and explore them in future studies.

\section*{Ethics Statement}

We address the efficiency of data annotation by investigating learning curves to estimate the necessary training sample size to reach a desired model performance. However, it is imperative to take into consideration the potential biases that may exist in the model predictions when utilizing a reduced amount of labeled data in the system construction process. Furthermore, when addressing complex tasks such as machine translation and text summarization, it is essential to guarantee the factuality of output generated by the system trained with the suggested data sample size.


%% file: sections/appendix.tex
\section{Detailed Results}
\label{appendix_detailed_results}

\subsection{Predicting the Required Data Size}
Table \ref{table:predicting_100_pct_performance} presents the results of required data size prediction using threshold $\alpha=0.1$ and $\alpha=0.3$.

\begin{table}[ht]
\centering
\resizebox{\columnwidth}{!}{
\centering
\begin{tabular}{lccccc}
\Xhline{2\arrayrulewidth}
\textsc{Benchmark} & \textsc{Cls (\#)} & \textsc{Size (\%)} & \textsc{Size (\#N)}  & \textsc{L1}$\downarrow$  & \textsc{100\%} \\\hline
\multicolumn{6}{c}{\textsc{$\alpha=0.1$}}\\\hline
\textsc{IMDb} & $2$ & $19$\% &  $16,800$ & $6.56$ & $17$K \\ 
\textsc{SST2} &$2$ & $8$\%  & $25,458$ & $8.27$ & $47$K \\
\textsc{AG News}  & $4$ & $28$\% & $82,320,$ & $2.96$ & $84$K \\
\textsc{DBpedia}  & $14$ & $27$\% & $384,160$ & $3.44$ & $392$K \\

\hline
\multicolumn{6}{c}{\textsc{$\alpha=0.3$}}\\\hline
\textsc{IMDb}  & $2$ & $96$\% &  $3,325$ & $5.84$ & $17$K \\ 
\textsc{SST2} & $2$ & $54$\%  & $3,772$ & $0.704$ & $47$K \\
\textsc{AG News}  & $4$ & $98$\% & $23,521$ & $9.9$ & $84$K \\
\textsc{DBpedia}  & $14$ & $98$\% & $105,840$ & $9.68$ & $392$K \\

\Xhline{2\arrayrulewidth}
\end{tabular}}
\caption{\small We show the results with the thresholds $\alpha=0.1$ and $\alpha=0.3$. \textsc{Size (\%)} stands for the percentages of the data size for the learning curve modeling, \textsc{Size (\#N)} is the number of the corresponding data size, \textsc{L1} is the L1 distance between the accuracy of models using all the training data and the estimated accuracy based on the saturated point. \textsc{100\%} specifies all training samples.}
\label{table:predicting_100_pct_performance}
\end{table}

\subsection{Data Weighting}
\label{sec:data_weighting}
We apply data weighting on three extrapolating functions using gradient decent methods in \ref{table:weighting_data_size_gd}.

\begin{table}[h]
\centering
\resizebox{\columnwidth}{!}{
\centering
\begin{tabular}{lcc}
\Xhline{2\arrayrulewidth}
\multirow{2}{*}{\textsc{Extrapolating}} & \multicolumn{2}{c}{\textsc{Gradient Descent}} \\
\cline{2-3}
& \textsc{Unweighted} & \textsc{Weighted} \\
\hline
\textsc{Exp}  & $0.0417$ &  $\textbf{0.0342}$  \\ 
\textsc{Inv} & $0.0706$  & $\textbf{0.0519}$ \\
\textsc{Pow4}  & $0.0979$ & $\textbf{0.0652}$ \\

\Xhline{2\arrayrulewidth}
\end{tabular}}
\caption{Better curve fitting when weighting data points at latter phase. We examine the effectiveness of weighting data size on the exponential (\textsc{Exp}), inverse power law (\textsc{Inv}), power4 (\textsc{Pow4}) function using gradient decent method. The learning curves fit on 5\%, 10\%, 25\% and 50\% data sizes of IMDB and is evaluated on testing sample with mean absolute error (MAE).}
\label{table:weighting_data_size_gd}
\end{table}

\begin{figure}[ht]
    \resizebox{\columnwidth}{!}{
    \centering
    \includegraphics[width=\textwidth]{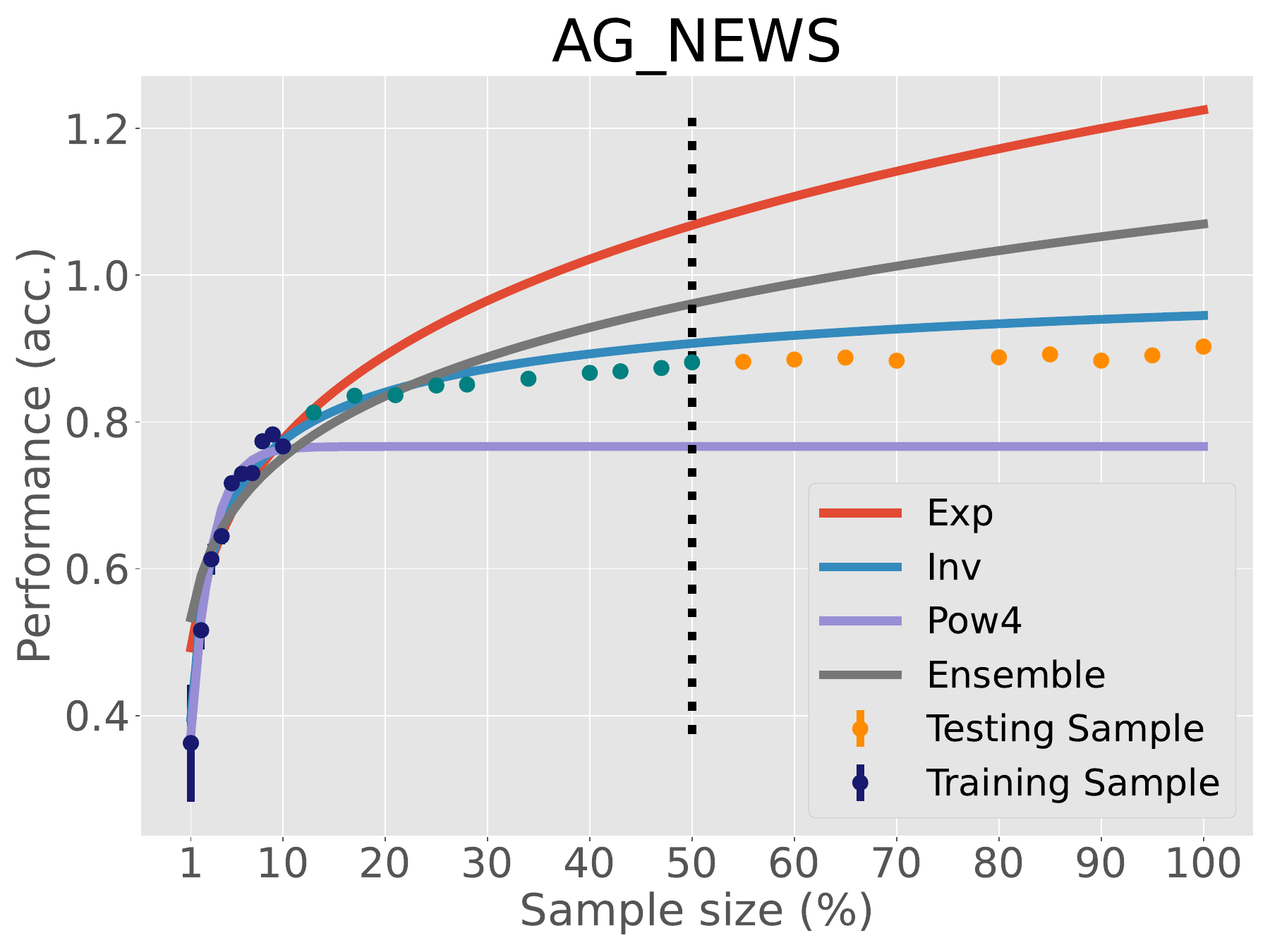}
    }
    \caption{Learning curves on $10\%$ data size using \textsc{AG NEWS}.    }
    \label{fig:ag_news_10}
\end{figure}

\subsection{Learning curve on 10\% data sizes of AG NEWS }\label{lr_on_ag_news}
Figure \ref{fig:ag_news_10} shows the learning curves fitting on $10$\% data sizes of \textsc{AG NEWS} dataset.

\begin{figure}[ht]
    \resizebox{\columnwidth}{!}{
    \centering
    \includegraphics[width=\textwidth]{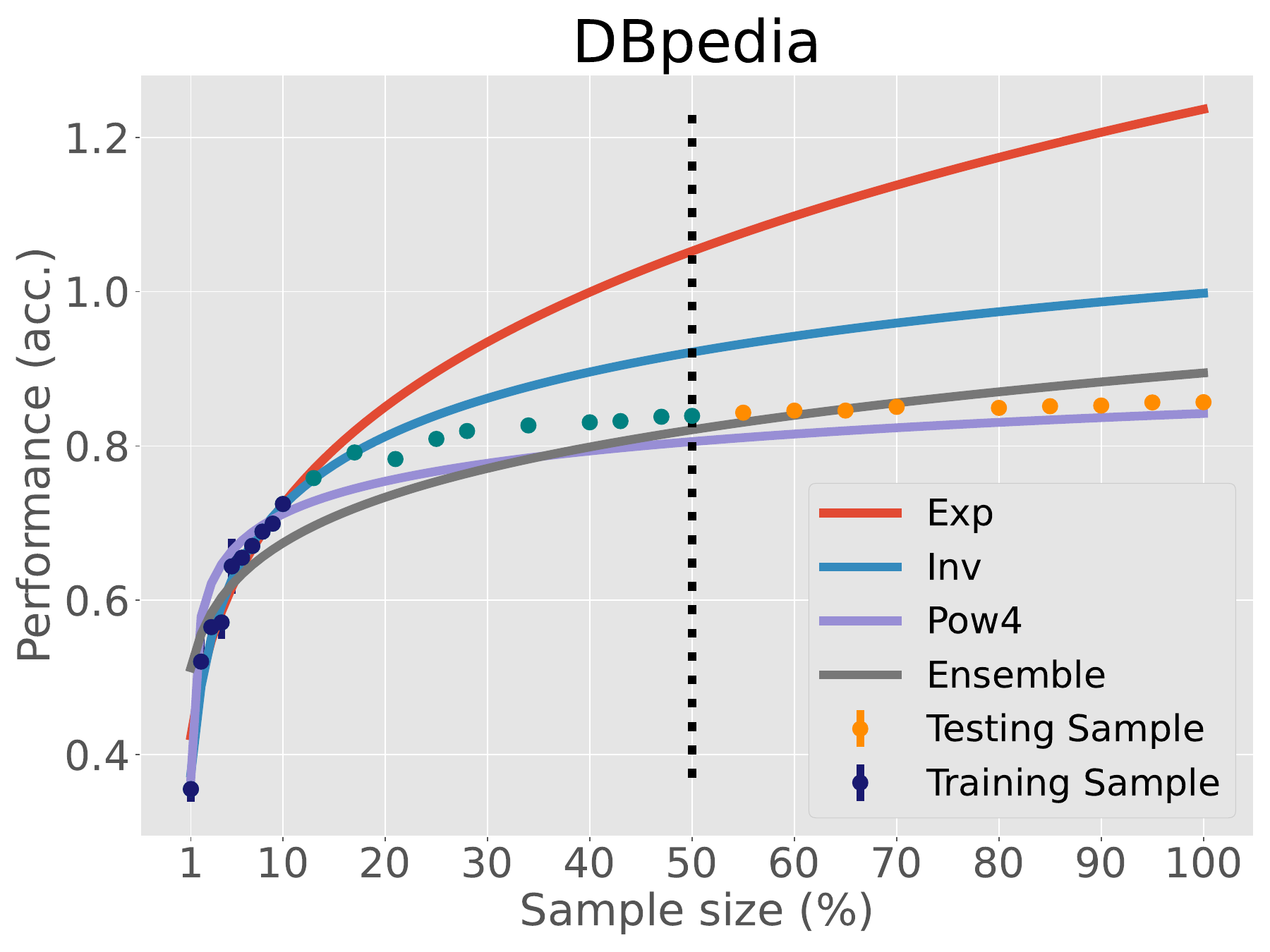}}
    \caption{Learning curves on $10\%$ data size using \textsc{DBpedia}.
    }
    \label{fig:dbpedia_10}
\end{figure}

\subsection{Learning curve on 10\% data sizes of DBpedia}
Figure \ref{fig:dbpedia_10} shows the learning curves fitting on $10$\% data sizes of \textsc{DBpedia} dataset.

\subsection{Effect of Sample Sizes for Learning Curve Fitting}\label{sec:sample_sizes}
We examined the relationship between sample sizes and the difference in mean absolute error (MAE)  between the predicted and actual performance trends across four benchmarks. Table \ref{table:data_sizes} showed MAEs when \textsc{Ensemble} fitting on $50\%$ and $10\%$ of data respectively. We observed that having more samples does not necessarily lead to a better model and that using only 10\% resulted in smaller MAEs on all four benchmarks. Therefore, we select 10\% of data points for learning curve modeling.

\begin{table}[h]
\centering
\resizebox{0.8\columnwidth}{!}{
\centering
\begin{tabular}{lcc}
\Xhline{2\arrayrulewidth}
\multirow{2}{*}{\textsc{Benchmark}} & \multicolumn{2}{c}{\textsc{Sample Sizes}} \\
\cline{2-3}
& \textsc{50\%} & \textsc{10\%} \\
\hline
\textsc{IMDb}  &  $0.0458$ &  $\textbf{0.00961}$ \\
\textsc{SST2} & $0.0299$ & $\textbf{0.0132}$  \\
\textsc{AG News}  & $0.0704$ & $\textbf{0.0209}$ \\
\textsc{DBpedia}  & $0.0734$ & $\textbf{0.0158}$  \\
\Xhline{2\arrayrulewidth}
\end{tabular}}
\caption{\small Learning Curve Fitting on $50\%$ and $10\%$ data size respectively.}
\label{table:data_sizes}
\end{table}